\pdfoutput=1

\documentclass[11pt]{article}

\usepackage[]{acl}

\usepackage{times}
\usepackage{latexsym}
\usepackage{graphicx}
\usepackage[T1]{fontenc}

\usepackage[utf8]{inputenc}

\usepackage{microtype}
\usepackage{array,booktabs,longtable,tabularx,tabulary}

\title{UTNLP at SemEval-2022 Task 6: A Comparative Analysis of Sarcasm
Detection Using Generative-based and Mutation-based Data Augmentation}

\newcommand*\samethanks[1][\value{footnote}]{\footnotemark[#1]}

\author{Amirhossein Abaskohi, Arash Rasouli\thanks{equal contribution}, Tanin Zeraati\samethanks, Behnam Bahrak \\
        School of Electrical and Computer Engineering, College of Engineering \\ University of Tehran, Tehran, Iran \\ \texttt{\{amir.abaskohi, arash.rasouli, t.zeraati, bahrak\}@ut.ac.ir}}

\begin{document}
\maketitle
\begin{abstract}
Sarcasm is a term that refers to the use of words to mock, irritate, or amuse someone. It is commonly used on social media. The metaphorical and creative nature of sarcasm presents a significant difficulty for sentiment analysis systems based on affective computing. The methodology and results of our team, UTNLP, in the SemEval-2022 shared task 6 on sarcasm detection are presented in this paper. We put different models, and data augmentation approaches to the test and report on which one works best. The tests begin with traditional machine learning models and progress to transformer-based and attention-based models. We employed data augmentation based on data mutation and data generation. Using RoBERTa and mutation-based data augmentation, our best approach achieved an F1-sarcastic of 0.38 in the competition's evaluation phase. After the competition, we fixed our model's flaws and achieved an F1-sarcastic of 0.414. 
\end{abstract}

\section{Introduction}

Billions of internet users use social networks not only to stay in touch with friends, meet new people, and share user-generated content but also to express their opinions on a wide range of topics using a variety of methods such as posting comments, videos, photos, etc. with specific groups of people \cite{tungthamthiti2016recognition}. In these platforms, users could submit information on whatever topic they wanted, with no restrictions on the sort of content they may share. The lack of constraints and individuals' anonymity on these networks led to humorous sarcastic texts.

Because sarcasm indicates sentiment, detecting sarcasm in a text is critical for anticipating the text's accurate sentiment, making sarcasm detection a valuable tool with multiple applications in domains such as security, health, services, product evaluations, and sales. Sarcasm detection is an essential aspect of creative language comprehension \cite{veale2019systematizing} and online opinion mining \cite{kannangara2018mining}. Even for humans, identifying sarcasm is difficult due to heavily contextualized expressions \cite{walker2012corpus}. There are few labeled data resources for sarcasm detection. Any available texts that can be collected (for example, Tweets) contain many issues, such as an evolving dictionary of slang words and abbreviations, requiring many hours of human annotation to prepare the data for any potential use. Furthermore, the nature of sarcasm identification adds to the task's difficulty, as sarcasm may be considered relative and varies significantly across people, depending on a variety of criteria such as the context, area, time, and events surrounding the statement.

In an attempt to solve this issue, we participated in SemEval-2022 shared task 6 \cite{abufarha-etal-2022-semeval}, which aims to recognize whether a tweet is sarcastic or not. Our contributions are as follows:
\begin{enumerate}
  \itemsep0em 
  \item {We experiment with simple machine learning models like Support Vector Machine (SVM) and various word encodings. }
  \item {To discover the optimum data preprocessing method, we tested the effect of various data preprocessing. }
  \item {We put several data augmentation techniques to the test. }
  \item {On our best dataset, we evaluated Long Short Term Memory (LSTM) based models, Bidirectional Encoder Representations from Transformers (BERT) based models, and attention-based models. Different neural network topologies are compared, and the model with the highest performance is reported.}
\end{enumerate}
With RoBERTa (A Robustly Optimized BERT Pretraining Approach), no preprocessing, and mutation-based data augmentation, our top result gets an F1-sarcastic of 0.38. However, we obtain better outcomes, with a 0.414 F1-sarcastic after fixing the problems of our proposed method.

The rest of this paper is organized as follows. In Section \ref{section:related}, we discuss the related work, Section \ref{section:dataset} introduces the dataset. In Sections \ref{section:method} and \ref{section:result}, we present our methodology and results, respectively. Finally Section 6 concludes the paper.

\section{Related Work}
\label{section:related}

We give a quick review of previous works on sarcasm detection in this part, followed by works on data augmentation.

\subsection{Sarcasm Detection on Twitter}
Sarcasm detection has been represented as a binary classification issue, with most tweets labeled with specific hashtags (e.g., \#sarcasm, \#sarcastic) being considered sarcastic. Many techniques in various languages have been proposed using this framework.

In \cite{davidov2010semi}, Semi-supervised sarcasm detection experiments were done using a Twitter dataset (5.9 million tweets) and 66,000 Amazon product evaluations. On the product review dataset, they acquired an F-measure of 0.83. On the Twitter dataset, they obtained an F-measure of 0.55 using 5-fold cross-validation on their k-Nearest Neighbor (kNN) like classifier.

\cite{gonzalez2011identifying} used 900 messages from Twitter sorted into three groups (sarcastic, positive sentiment, and negative sentiment). To find sarcastic tweets, they utilized the hashtags \#sarcasm and \#sarcastic. SVM with Sequential Minimum Optimization (SMO) and logistic regression were employed as classifiers. The best accuracy for the sarcastic class was 0.65.

\cite{reyes2012humor} presented elements to capture ambiguity, polarity, unexpectedness, and emotive situations in figurative language. F1-sarcastic of 0.65 was the best result in categorizing irony and general tweets.

The representativeness and significance of conceptual elements have been investigated in \cite{reyes2013multidimensional}. Punctuation marks, emoticons, quotations, capitalized words, lexicon-based features, character n-grams, skip-grams, and polarity skip-grams are all examples of these characteristics. Each of the four categories (irony, comedy, education, and politics) in their corpus has 10,000 tweets. Using the Naive Bayes and decision trees algorithms, they evaluated two distributional scenarios: balanced distribution and unbalanced distribution (25\% ironic tweets and 75\% tweets from the three non-ironic categories). The decision trees classified the balanced distribution with an F1-sarcastic of 0.72 and the unbalanced distribution with an F1-sarcastic of 0.53.

One sort of sarcasm identified by \cite{riloff2013sarcasm} is the difference between a good mood and a bad scenario. Using a bootstrapping approach, the authors gathered collections of positive sentiment phrases and negative circumstance words from sarcastic tweets. They suggested a method for classifying tweets as sarcastic if they contain a positive predictive close to a negative context phrase. They used a SVM classifier using unigrams and bigrams as features to evaluate a human-annotated dataset of 3000 tweets (23\% sarcastic), getting an F1-sarcastic of 0.48. The F1-sarcastic of the hybrid strategy, which combined the findings of the SVM classifier with their baseline method, was 0.51.

\cite{lukin2017really} used bootstrapping, syntactic patterns, and a high precision classifier to classify sarcasm and nastiness in online chats. On their snark dataset, they got an F1-sarcastic of 0.57.

In \cite{oprea2019isarcasm}, LSTM, Att-LSTM, CNN, SIARN, MIARN, 3CNN, and Dense-LSTM models were used to assess the task dataset that was introduced in \cite{oprea2019isarcasm}, which is an unbalanced dataset and labeled by the tweets' writers. Using Multi-Dimension Intra-Attention (MIARN) \cite{tay2018reasoning} Network, they could get an F-score of 0.364.

In \cite{guo2021latent}, the Latent Optimized Adversarial Neural Transfer (LOANT) model was suggested as a novel latent-optimized adversarial neural transfer model for cross-domain sarcasm detection. LOANT surpasses classical adversarial neural transfer, multitask learning, and meta-learning baselines using stochastic gradient descent (SGD) with a one-step look-ahead and sets a new state-of-the-art F-score of 0.4101 on the iSarcasm dataset.

\subsection{Data Augmentation}
Natural Language Processing(NLP) encompasses a wide range of tasks, from text categorization to question answering, but no matter what you do, the quantity of data you have to train your model has a significant influence on the model's performance. Using the data you already have, data augmentation techniques are used to produce extra, synthetic data. Augmentation techniques are widely used in computer vision applications, but they may also be used in natural language processing.

In the instance of Twitter, \cite{van2018semeval} and \cite{ilic2018deep} found that adding more data from the same domain did not improve the performance for recognizing sarcasm and irony. Although their result is not general for all sarcasm detection tasks and the result of data augmentation depends on the data and augmentation method.

\cite{lee2020augmenting}'s idea is to make a new datapoint out of the context sequence [c1, c2,, cn] and label it "NOT SARCASM." The sequence could not be identified as "SARCASM" without the answer [r1]. They believe that the newly created negative samples will aid the model in focusing on the link between the response [r1] and its contexts [c1, c2, cn]. They also create positive samples using back-translation procedures\cite{berard2019naver, zheng2019robust} in French, Spanish, and Dutch to balance out the quantity of negative examples.

In \cite{feng2020genaug} different data augmentation methods were tested on Yelp Reviews dataset\cite{Yelpp} for GPT-2 generative model\cite{radford2019language}. They used "Random Insertion, Deletion, \& Swap", "Semantic Text Exchange (STE)", "Synthetic Noise", and "Keyword Replacement". They showed in some case data  augmentation could help them to reach better performance.

This paper is the first to look at generative-based and mutation-based data augmentation strategies in sarcasm detection.

\section{Dataset}
\label{section:dataset}

We mostly used the iSarcasm \cite{oprea2019isarcasm} dataset in this study. In specific experiments, we integrated the primary dataset with various secondary datasets, including the Sarcasm Headlines Dataset \cite{misra2019sarcasm} and Sentiment140 dataset \cite{go2009twitter} to increase the quantity of data and compensate for the lack of sarcastic data. For each dataset, the details are further discussed. It is worthy to mention that all of the supplementary datasets we included had a negative impact on our model's performance. We believe this was the result of a different data gathering method. Because to the differing labeling process and domain, the distribution diverged from that of iSarcasm. As a result, the following sections are solely dependent on the iSarcasm dataset, with no other datasets being used.

\subsection{Main Task Dataset: iSarcasm}
According to \cite{oprea2019isarcasm}, the sarcasm labeling using hashtags to build datasets captures just the sarcasm that the annotators were able to detect, leaving out the intended sarcasm. When the author intends for the content to be sarcastic, it is called intended sarcasm. The iSarcasm dataset includes 4484 tweets: 3707 non-sarcastic and 777 sarcastic. Because some tweets had been erased, we only had access to 3469 tweets for the job. The unbalanced dataset and the scarcity of sarcastic data were two of the most significant issues we encountered. Table \ref{table: Main Dataset} displays some of the dataset's annotated remarks.

\begin{table*}
  \caption{Example of Sarcastic and Non-Sarcastic tweets.}
  \label{table: Main Dataset}
  \begin{center}
  \begin{tabular}{cccl}
    \toprule
    Tweet & Sarcastic & Sarcasm Type \\
    \midrule
    \texttt Oh my goodness. It's the first week of the &  &   \\
    \texttt summer holidays and @name has found & Sarcastic &  ['Sarcasm'] \\
    \texttt his recorder Give.Me.Strength. &  &   \\
    \hline
    \texttt 90\% of adulthood is just refilling your @name pitcher. & Sarcastic & ['Irony', 'overstatement'] \\
    \hline
    \texttt True bliss is laying in an ice\\
    cold bath during the hottest part of the year & Non-Sarcastic & []\\
    \bottomrule
  \end{tabular}
  \end{center}
\end{table*}

\subsection{Sarcasm Headlines Dataset}
Sarcasm Headlines Dataset \cite{misra2019sarcasm, misra2021sculpting} was gathered from two news websites. It is beneficial since it overcomes the constraints of Twitter datasets due to noise. As the second edition of this dataset includes more data and a greater variety of data than the first version, we chose the second version.

\subsection{Sentiment140 Dataset}
We needed to compensate for the limited data to train our model successfully. As a result, we chose the sentiment140 dataset \cite{go2009twitter} because it has a large quantity of data and is based on Twitter. The sentiment tweet message is labeled using an automated classification approach in this dataset. The accuracy is more than 80\% when using a machine learning algorithm.

\section{Methodology}
\label{section:method}

In this study we examined and analyzed various models and data augmentation strategies for sarcasm detection. First, we go through data augmentation methods; then, we discuss the structure and hyperparameters of these models in this section. The codes of all models are  available on GitHub\footnote{\href{https://github.com/AmirAbaskohi/SemEval2022-Task6-Sarcasm-Detection}{https://github.com/AmirAbaskohi/SemEval2022-Task6-Sarcasm-Detection} }.

\subsection{Data Augmentation}

\subsubsection{Generator-based}
For this augmentation method, we used GPT-2 \cite{radford2019language} generative model to generate 4000 tweets for both sarcastic and non-sarcastic classes. Then we selected 2000 tweets of each class randomly to increase dataset quantity and have more sarcastic samples.

\subsubsection{Mutation-based}
\begin{figure}
  \centering
  \includegraphics[width=7cm,height=5cm,keepaspectratio]{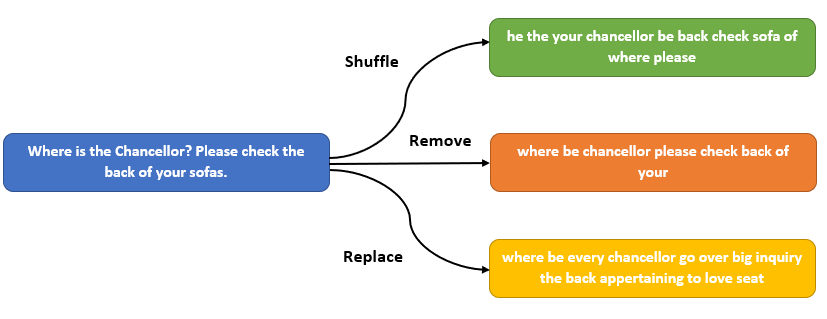}
  \caption{Effect of shuffling, word elimination, and replacing with synonyms on a tweet sample.}
  \label{fig:Mutation-based}
\end{figure}

We used three distinct ways to change the data in this method: eliminating, replacing with synonyms, and shuffling. These processes were used in the following order: shuffling, deleting, and replacing. The removal and replacement were carried out systematically. We used the words' roots to create a synonym dictionary. Synonym dictionary is created by scarping the Thesaurus website\footnote{\href{https://www.thesaurus.com}{https://www.thesaurus.com}}. When a term was chosen to be swapped with its synonyms, we chose one of the synonyms randomly (Figure \ref{fig:Mutation-based}). We tried each combination of these processes to find the best data augmentation combination (a total of seven).

\subsection{Models}
\subsubsection{Support Vector Machine (SVM)}
We utilized SVM to discover the optimal approaches for dataset preprocessing and word embeddings. For data augmentation, we employed both generator-based and mutation-based methods. We also put other data preprocessing approaches to the test, such as link removal, emoji removal, stop word removal, stemming, and lemmatizing. We utilized TF-IDF, Word2Vec \cite{mikolov2013efficient}, and BERT \cite{devlin2018bert} for word embedding. We found that using a regularization value of 10 and a Radial Basis Function (RBF) kernel, BERT word embedding, and no data preprocessing will give us the best results.

\subsubsection{LSTM-based Methods}
We begin with the intuition that a memory model can help us reach a better result. So we started with Long Short Term Memory (LSTM) model \cite{hochreiter1997long}. We used one LSTM layer followed by time distributed dense layer. We repeated these two layers one more time, and then we used another LSTM layer followed by two dense layers. This model and all of the following models in this section were trained in 10 epochs. 

In addition, we used Bidirectional Long Short Term Memory (BLSTM). Using bidirectional will run the inputs in two directions, one from past to future and the other from future to past. We used one BLSTM layer for this network, followed by a time-distributed dense layer. We repeated these two layers one more time, and then we used another BLSTM layer followed by two dense layers.

Furthermore, we combined LSTM and BLSTM with Convolutional Neural Networks (CNNs). CNN layers for feature extraction on input data are paired with LSTM to facilitate sequence prediction in the CNN-LSTM architecture.
Although this model is often employed for video datasets, \cite{rehman2019hybrid}  demonstrated that it could perform better in sentiment analysis tasks.
We used three 1D convolutional layers followed by a 1D global max-pooling layer for the convolutional part. We used these layers at the end of LSTM-based networks.

\subsubsection{BERT-based Methods}
The use of bidirectional training of transformer and a prominent attention mode for language modeling is BERT's fundamental technological breakthrough \cite{devlin2018bert}. The researchers describe a new Masked Language Model (MLM) approach that permits bidirectional training in previously tricky models.
They found that bidirectionally trained language models can have a better understanding of language context and flow than unidirectional ones. 

Robustly Optimized BERT or RoBERTa has a nearly identical architecture to BERT, however, the researchers made some minor adjustments to its architecture and training technique to enhance the results on BERT architecture \cite{liu2019roberta}.

We used both RoBERTa with twitter-roberta-base, which has been trained on near 58 million tweets and finetuned for sentiment analysis with the TweetEval benchmark and BERT with bert-base from Huggingface \cite{wolf2019huggingface}. For both models, we employed five epochs, batch size of 32, 500 warmup steps, and a weight decay of 0.01.

\subsubsection{Attention-based Methods}

One of the most important achievements in deep learning research in the recent decade is the attention mechanism \cite{vaswani2017attention}. The Encoder-Decoder model's restriction of encoding the input sequence to one fixed-length vector to decode each output time step is addressed via an attention mechanism. This difficulty is thought to be more prevalent when decoding extended sequences.

We start with the assumption that if a model with an attention layer is trained to identify sarcasm at the sentence level, the sarcastic words will be the ones the attention layer learns to value. As a result, we added an attention layer to our LSTM-based and BERT-based models. The results will be discussed further.

\subsubsection{Google's T5}

Google's T5 \cite{raffel2019exploring} text-to-text model outperformed the human baseline on the GLUE, SQuAD, and CNN/Daily Mail datasets and earned a remarkable 88.9 on the SuperGLUE language benchmark.

\begin{figure}
  \centering
  \includegraphics[width=7cm,height=5cm,keepaspectratio]{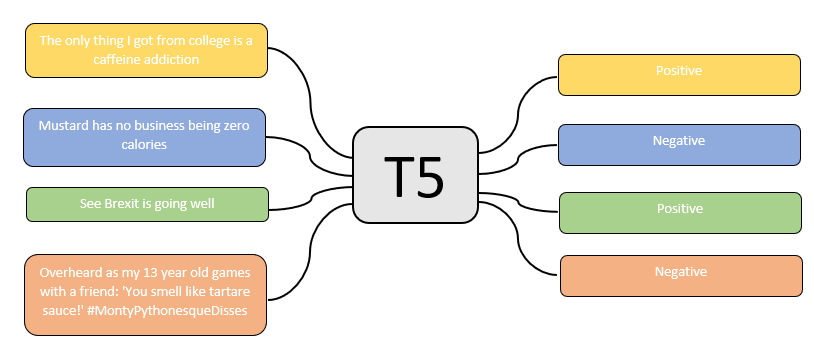}
  \caption{Fine-tuning T5 model for sarcasm detection problem.}
  \label{fig:T5}
\end{figure}

We fine-tuned T5 for our problem and dataset by giving the sarcastic label the target and the tweets as the source. We used two epochs, batch size of 4, 512 tokenization max length, Adam epsilon of 1e-8, word decay of 0, no warmup steps, and learning rate of 3e-4 (Figure \ref{fig:T5})\footnote{We were not able to test a larger version of the model with better hyperparameters due to resource constraints}.

\section{Results}
\label{section:result}
In this section we report the results of our models introduced in Section \ref{section:method}.

It's important to note that after the competition, we discovered that none of our preprocessing strategies improved the performance of our model. So we were able to get an F1-sarcastic of 0.414 without using any preprocessing methods, which was 0.034 higher than our performance in the competition, which was based on the best combination of preprocessing methods.

\begin{table}
  \caption{F1-sarcastic and accuracy for different data augmentation methods on SVM model with BERT word embedding and no preprocessing.}
  \label{table: AugRes}
  \begin{center}
  \small
  \begin{tabular}{ccc}
    \toprule
    Data Augmentation & F1-sarcastic & Accuracy\\
    \midrule
    Shuffling & 0.305 & 0.7471\\
    \hline
    Shuffling + Replacing & 0.301 & 0.741\\
    \hline
    Shuffling + Removing & 0.306 & 0.747\\
    \hline
    Removing & 0.301 & 0.747\\
    \hline
    GPT-2 & 0.292 & 0.675\\
    \bottomrule
  \end{tabular}
  \end{center}
\end{table}

\subsection{Support Vector Machine (SVM)}
The optimum augmentation technique, preprocessing method, and word embedding were all determined using the SVM model. Without any augmentation, BERT obtained the greatest F1-sarcastic of 0.2862, compared to 0.2541 and 0.0924 for Word2Vec and TF-IDF, respectively.

We have also looked at several ways of data augmentation. The F1-sarcastics for shuffling with replacing words, only word elimination, just shuffling, and shuffling with word elimination were the highest in the mutation-based augmentation (Table \ref{table: AugRes}). We also tried these data augmentation and GPT-2 data augmentation on RoBERTa because the results were close, and we found that merely word removal was the best data augmentation. The following results are based on no data preprocessing, BERT word embedding, and mutation-based data augmentation utilizing only word removal.  

\subsection{LSTM-based Methods}
 LSTM obtained an F1-sarcastic of 0.2176 using BERT word embeddings, mutation-based data augmentation, and no preprocessing, whereas BLSTM's F1-sarcastic was 0.2439 using BERT word embeddings, mutation-based data augmentation, and no preprocessing. By adding CNN layers, the F1-sarcastic of the LSTM was increased to 0.2453, and the BLSTM was increased to 0.2751. The CNN model's F1-sarcastic was 0.2263.

\subsection{BERT-based Methods}
We employed a mutation-based data augmentation approach with no preprocessing for BERT-based procedures. We got an F1-sarcastic of 0.323 using BERT. We achieved our best result with RoBERTa with an F1-sarcastic of 0.414, which was better than LOANT \cite{guo2021latent} model on the same dataset.


\begin{table}
  \caption{Best results for each model using iSarcasm dataset and mutation-based data augmentation.}
  \label{table: Res}
  \begin{center}
  \begin{tabular}{ccc}
    \toprule
    Model & F1-sarcastic & Accuracy\\
    \midrule
    SVM & 0.3064 & 0.7478\\
    \hline
    LSTM-based & 0.2751 & 0.7251\\
    \hline
    BERT-based & 0.414 & 0.8634\\
    \hline
    Attention-based & 0.2959 & 0.7793\\
    \hline
    Google's T5 & 0.4038 & 0.8124\\
    \bottomrule
  \end{tabular}
  \end{center}
\end{table}

\subsection{Attention-based Methods}
Adding attention layers to this job was not helpful, and it decreased our models' performance. RoBERTa's F1-sarcastic dropped to 0.2959 using the attention layer. LSTM model with the attention layer earned an F1-sarcastic of 0.2145. The F1-sarcastic of BLSTM with attention layer was 0.2336.

\subsection{Google's T5}
Based on the hyperparameters listed in the Section \ref{section:method}, our F1-sarcastic for this model is 0.4038. However, we believe that we may get better results by increasing the tokenization max length, increasing the batch size, and utilizing the t5-large pre-trained model.

\section{Conclusion}
In this study, we reviewed and contrasted a number of sarcasm detection methods. To improve the performance of our model, we experimented with two different types of augmentation. In the job of sarcasm detection, we observed that mutation-based data augmentation can assist us in achieving better results than generative-based data augmentation. Additionally, we tested with other deep-learning techniques, including RNN and BERT-based models. Our best system, an ensemble model, has an F1-sarcastic of 0.414.

\section*{Acknowledgements}
We want to convey our heartfelt gratitude to Prof. Yadollah Yaghoobzadeh, who provided us with invaluable advice during our research.

\bibliography{anthology,custom}

\appendix



\end{document}